\documentclass{article}

     \usepackage[final]{neurips_2019}

\usepackage[utf8]{inputenc} 
\usepackage[T1]{fontenc}    
\usepackage{hyperref}       
\usepackage{url}            
\usepackage{booktabs}       
\usepackage{amsfonts}       
\usepackage{nicefrac}       
\usepackage{microtype}      
\usepackage[pdftex]{graphicx}

\usepackage{mathtools, nccmath}
\usepackage{float}

\title{Recurrent Autoencoder with Skip Connections and Exogenous Variables for Traffic Forecasting}

\author{%
  Pedro Herruzo \\
  Dept. of Computer Architecture\\
  Polytechnic University of Catalonia\\
  North Campus C6, 08034 Barcelona, Spain\\
  \texttt{pherruzo@ac.upc.edu} 
  \And
  Josep L. Larriba-Pey \\
  Dept. of Computer Architecture\\
  Polytechnic University of Catalonia\\
  North Campus C6, 08034 Barcelona, Spain\\
  \texttt{larri@ac.upc.edu} \\
}

\begin{document}

\maketitle

\begin{abstract}
The increasing complexity of mobility plus the growing population in cities, together with the importance of privacy when sharing data from vehicles or any device, makes traffic forecasting that uses data from infrastructure and citizens an open and challenging task. In this paper, we introduce a novel approach to deal with predictions of speed, volume, and main traffic direction, in a new aggregated way of traffic data presented as videos. The approach leverages the continuity in a sequence of frames and its dynamics, learning to predict changing areas in a low dimensional space and then, recovering static features when reconstructing the original space. Exogenous variables like weather, time and calendar are also added in the model. Furthermore, we introduce a novel sampling approach for sequences that ensures diversity when creating batches, running in parallel to the optimization process.
  
\end{abstract}

%
%



\section{Introduction}

Traffic forecasting deals with the problem of regression and classification of road states taking into account Spatio-temporal features. Road states refers to traffic variables \cite{RESPATI2018131} \textit{speed} (average speed of cars at a specific location or area), \textit{volume} (number of cars in this location), and \textit{direction} (the angle from 0 to 359 along which vehicles move). This problem, consisting of assigning future road states at each location, is particularly difficult mainly due to (i) the complexity and dynamic property of the traffic environment in cities \cite{Liao2018MultimodalST}, (ii) high resolution of the data (iii) lack of up-to-date road maps and data in some locations, (iv) external impacts of unknown agents to traffic like weather or soccer games, and (v) errors in data collection from sensors.


The current complexity of mobility, the growing population in cities and the increasing traffic-data collection \cite{RESPATI2018131} (Loop detectors, Bluetooth Mac Scanners, Mobile Phones or Connected Cars), calls for more powerful models that allow better city planning and calculating more precisely travel times. In this regard, Traffic4cast Challenge 2019 at NeurIPS \cite{IARAI} proposes a new representation of traffic data that deliberately ignores the underlying road network, mapping GPS trajectories to Spatio-temporal cells that do not depend on the lack of up-to-date road maps, which empowers the development of methods that can produce high-resolution traffic states even for fast-evolving cities, like emerging economies.

Traffic forecasting is traditionally \cite{SVR} treated as a time series problem faced with methodologies like Arima, Sarima or Support Vector Regression while more recent works use also Neural networks \cite{6786503}. However, these methods take into account the state of streets independently while there is a strong correlation between streets in real life. After the great success of Convolutional Neural Networks (CNN) in Computer Vision with images, or with tasks related to Signal Processing or Natural Language, the work in \cite{8258162} translates traffic data into images and uses for the first time a CNN to tackle this problem. The most relevant publication for our work is Hierarchical Long-term Video Prediction without Supervision \cite{Wichers2018HierarchicalLV}, where the prediction of a future frame is done using an encoder-decoder scheme, with predictions occurring in the embedded space and then gradually recovering back the original dimension with the decoder, all minimizing both the output of the next frame and the prediction in the embedding space.

This paper presents a novel method to tackle the problem of traffic forecasting as a scene completion task along time using the data provided by the Institute of Advanced Research in Artificial Intelligence (IARAI) \cite{IARAI} together with Here Technologies in the NeurIPS Traffic4cast challenge aforementioned. The proposed method is based on \cite{Wichers2018HierarchicalLV}, using their proposed loss function but reshaping the network as a sequence to sequence problem and using a U-Net \cite{RESPATI2018131} like architecture together with exogenous data like time of the day, day of the week and weather, benefiting from both the prediction in a lower dimensional space and the output of several future frames within a single inference. We also present a data sampling technique that allows for faster training and batch diversity. Furthermore, we propose a multitask loss function to minimize, taking into account that the regression for the heading channel for these data is indeed a classification problem.

The rest of the paper is organized as follows. In Section \ref{Problem_definition}, we describe the problem to be solved and the data. Then, in section \ref{Methodolody} we present the proposed methods for data sampling and frames prediction. Section \ref{Results} discusses the experimental results and section \ref{Conclusions} concludes the manuscript with remarks and future work.  

\section{Problem definition} \label{Problem_definition}
\begingroup
Traffic4cast presents the problem of traffic forecasting as a scene completion task along time for one year in three different cities: Berlin, Istanbul and Moscow. In particular, trajectories of raw GPS positions for each city are projected to an image containing the city with shape $\text{height}=495$,  $\text{width}=436$, and  $\text{channels}=3$ (\textit{speed}, \textit{volume} and \textit{heading} in this order). Each pixel in the image contains the aggregated information for a square region of $100m\times100m$ over a time bin of 5 minutes. As a day is composed of $288$ time bins of $5$ minutes, for each city and day we can represent the data as a tensor $T_{\textit{city}}^{\textit{day}}[t, h, w, c]$ with shape $(288, 495, 436, 3)$, with channel domains $\textit{speed} \in \{0, 1, 2, ..., 255\} \subset \mathbb{N}$, $\textit{volume} \in \{0, 1, 2, ..., 255\} \subset \mathbb{N}$, and $\textit{heading} \in \{0, 1,85,170,255\}$. The aggregation of \textit{volume} is the counting of vehicles in the interval and region $(t, h, w)$, capped at a minimum and maximum level to remove noisy readings. Then, capped numbers are mapped proportionally to the interval $[1, 255]$ and rounded to the nearest integer, where the value $0$ means no data available at this time bin. \textit{speed} is similarly calculated, only differing in the aggregation being the average instead of counting, and only capping the maximum speed and normalizing to the interval $[0, 255]$. Here $\textit{speed}=0$ means the vehicles are not moving, if \textit{volume} in this location is greater than zero. The computation of \textit{heading} is different, each probe point records the heading direction in degrees (from 0 to 359), which is binned in four heading directions; North-East (from 0 to 90, represented as $\textit{heading}=85$), South-East (from 90 to 180, as $\textit{heading}=255$), South-West (from 180 to 270, as $\textit{heading}=170$), and North-West (from 270 to 359, as $\textit{heading}=1$). The selected value is the bin with the highest number of points, with the possibility to assign 0 when all directions in this region have the same number of points, being impossible to determine which is maximum. Note that there is no data if and only if $\textit{volume}=0$. 


For each city Traffic4cast dataset provides 285 days for training, 7 for validation and 72 for testing. The first two have information for each time bin, but the test set only contains information in 5 blocks of 12 bins (1 hour of information each block). The goal of the challenge is to predict 3 time bins after each of these 5 blocks. In other words,  we need to predict the traffic for 5, 10 and 15 minutes ahead, given the information about the previous hour, a total of 5 times per day. In particular, the 5 blocks of 15 minutes to predict start in Istanbul and Moscow at time bins (57, 114, 174,222, 258), which correspond in hours to (04:45h, 09:30h, 14:30h, 18:30h, 21:30h) respectively. In Berlin, time bins are (30, 69, 126, 186, 234) that correspond to (02:30h, 05:45h, 10:30h, 15:30h, 19:30h).

\begingroup
\setlength\abovedisplayskip{1pt}
\setlength\belowdisplayskip{0pt}
Then, the problem can be formulated as follows. Given a \textit{$city$}, find a function $f$ such that: \begin{equation} 
\textit{f} = \min_{\widetilde{\textit{f}} \in \Theta} 
                \textit{L} \: ( \:
                    \widetilde{\textit{f}} \:
                        (\,T_{\textit{city}}^{\textit{day}}[s\!-\!q:s, h, w, c] \,),  \;
                    T_{\textit{city}}^{\textit{day}}[s:s\!+\!3, h, w, c] \:), 
            \forall \, \textit{day}, s, h, w, c 
\end{equation} where $s\!-\!q \geq 0$, $s\!+\!3 \leq 287$, $q \in \{1, 2, ...,  12\}$ is the length of the input sequence, $\textit{L}\:(\cdot,\cdot)$ is a \textit{loss function} that measures the error between the ground truth and the prediction, and $\Theta$ is the parameter space.
\endgroup

\section{Methodology} \label{Methodolody}

In this section, we present our twofold contribution; the sampling strategies of sequence to sequence data and our proposed models.

\subsection{Sampling strategies for sequences}
We define three sampling strategies that depend on the desired sequence input length $q \in \{1, 2, ...,  12\}$, while the output length is fixed to 3 frames. The first strategy (\textit{non overlapping}), consists in divide each day into $T_q = \textit{ceil}(288/(q+3))$ number of sequences without overlapping, where \textit{ceil} returns the smallest integer value greater than or equal to its input. For instance, with $q\!=\!3$, we can divide a day into $T_3\!=\!48$ sequences of length 6, 3 frames for training and 3 for testing. The second strategy (\textit{all possible slots}), makes use of every possible sequence of length $q\!+\!3$ starting from frame 0, then starting from frame 1, and so on until starting the last sequence from frame $288\!-\!(q\!+\!3\!-\!1)$. Note that with the example of $q\!=\!3$, method \textit{non overlapping} produces 285 days $\times$ 48 sequences $=13680$ sequences to train per city, whereas method \textit{all possible slots} produces $285 \times (288 - (3+3-1)) = 80655$ sequences, which is almost 6 times more sequences to train. The third strategy (\textit{like-test}), only trains sequences with the output time bins being the same as in the test set. Hence, in this procedure, we only have $285 \times 5 = 1425$ sequences to train. 

Each different strategy defines a set of time bins to train for each image. In order to impose diversity when creating each batch, our dataset is defined as a list of pairs (\textit{image}, \textit{time bin}) that are shuffled at the beginning of each epoch, creating batches that contain sequences of different days and time bins together. This should lead to faster convergence since batches are always different after each epoch, as explained by Yoshua Bengio in \cite{Bengio2012}. Furthermore, we perform all batch preparation and preprocessing in parallel to the optimization, which makes our training diverse and efficient. In preprocessing, we cast data to a float numbers and normalize all values into the interval [0, 1].


\subsection{Proposed models}

Our model builds upon the architecture proposed by Nevan Wichers et al. in \cite{Wichers2018HierarchicalLV}, where the authors presented a neural network that can predict the next frame ($\textit{F}_{t+1}$) in a movie given the previous one ($\textit{F}_t$). Given $\textit{F}_t$, an encoder followed by a recurrent layer predicts the embedding of the future frame, which is compared with $\textit{F}_{t+1}$ applied to the same encoder using an L2 loss. Then, the embedding is upsampled to the original space by a decoder, which is compared to $\textit{F}_{t+1}$ also with an L2 loss, using skip connections between different layers in the encoder-decoder to refine the output.

\begingroup
\setlength\abovedisplayskip{1pt}
\setlength\belowdisplayskip{0pt}
Given that the nature of our problem is a sequence to sequence prediction, we adapted the aforementioned architecture to take profit of the entire input sequence of length $q$ $\textit{X}_q$, when predicting the three future frames $\textit{Y}_3$. To do so, we accept a sequence of any size as input by iteratively using the encoder and concatenating its outputs. Then a recurrent encoder accumulates the temporal information of the input sequence into a single representation, and a recurrent decoder gives us three embedded predictions $\widetilde{\textit{e}_3}$. Afterward, these predictions are upsampled to the original space ($\widetilde{\textit{Y}_3}$) by a decoder that uses skip connections from each layer of the encoder, but only from the last frame in the input sequence including the image. As in the original model, we train both the predictions in the embedding and in the original space using an L2 loss with weights $\alpha, \beta \in (0, 1)$:

 \begin{equation} 
\textit{L} = \alpha \textit{L2}\,(\textit{Y}_3, \widetilde{\textit{Y}_3})  +  \beta \textit{L2}\,(\textit{e}_3, \widetilde{\textit{e}_3})
\end{equation}

Figure \ref{model_summary} illustrates our architecture. As can be seen, we further introduce exogenous data to our model. In particular, for each frame in the sequence, we introduce its time of day, day of the week, current weather and its prediction of the next 3 time bins using the predicted data provided by \cite{weather}. In our implementation, the encoder is composed of 6 blocks of i) two subblocks of (convolution, Batch normalization, and ReLu), ii) Max pooling, and iii) Dropout=0.5, with number of convolutions [16, 16x2, 16x4, 16x8, 16x8, 16x2] in this order for each block, and downsampling the image from 512 to 16 by powers of 2. The decoder is the same but using Transposed Convolutions to upsample the image instead of Max pooling. The recurrent encoder-decoder uses layers of GRU with (2048, 256, 128) and (128, 256, 2028) units respectively. Frames are upsampled using bilinear interpolation in the first place in the model to $512 \times 512$, and cropping plus a convolution of size 3 with ReLu are used at the end to match the original size $495 \times 436$.

In the rest of the paper, we will call the proposed model: Recurrent Autoencoder All (\textit{RAE\_all}), which includes skip connections (including the Input), weather and time information. We also tried different versions of this model: i) one without using the input of the last sequence in the skip connections for the decoder (\textit{RAE\_not\_In}), ii) another one that does not use exogenous variables (\textit{RAE\_not\_Exo}), and iii) a third one, were we explore a variant model (\textit{RAE\_Clf}) that outputs two regression channels for \textit{speed} and \textit{volume}, and five other outputs for the classification of the \textit{heading} channel, minimizing the first two with an L2 loss and the classification with a softmax crossentropy. Then, we use all 7 outputs to finally output the 3 original channels and minimizing again with an L2 loss. This way we believe that outputs in the \textit{heading} channel will be good for MSE but more accurate in the only 5 possible outputs thanks to the classification task.

We also define two baseline models. The first is called \textit{ConvLSTM} and consists of three layers of Convolutional LSTM with 32, 64, 64 units followed by the \textit{tanh} activation, and a final ConvLSTM layer with 3 units and a ReLu activation allowing to take the value of 1, needed for the \textit{heading} channel, all minimized with an L2 loss. The second variant is called \textit{ConvLSTM+Clf}, which adds an extra classification branch for the \textit{heading} channel making the model multitask, with a second loss being a softmax crossentropy.
\endgroup


\begin{figure}[H] \label{model_summary}
  \centering
  \includegraphics[width=1.\linewidth]{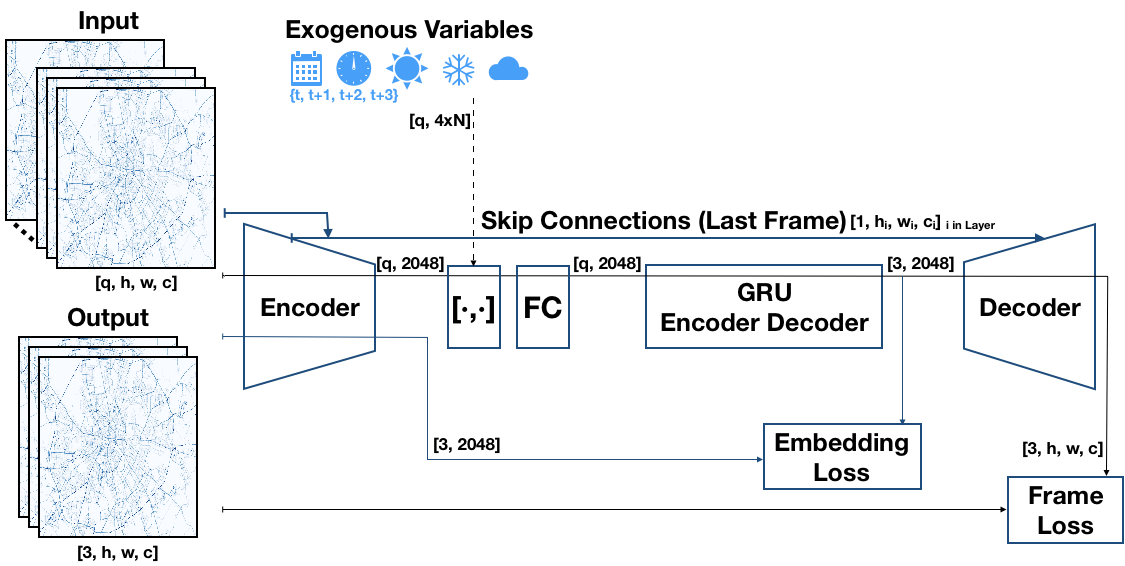}
  \caption{Recurrent Autoencoder with Skip Connections and Exogenous Variables. Embedding Loss makes the recurrent layer GRU together with the encoder to produce better predictions in a low dimensional space. Frames Loss and skip connections from sibling layers in the encoder empowers the decoder to produce outputs with high definition. Exogenous variables are concatenated with the encoder output plus a fully connected layer before recurrent layers. Best seen in electronic form. Our method code is available at \url{https://github.com/pherrusa7/Traffic4cast_NeurIPS_2019}.}
\end{figure}

\section{Results} \label{Results}
Table~\ref{tab:baselines} shows the results for baseline models in the three cities of the challenge. We do not only show the mean square error (mse) but also the accuracy for the \textit{heading} channel due to its intrinsic nature as a classification problem. As it can be seen, Moscow city presents very poor results comparing to Istanbul and Berlin. Baselines were only trained with the \textit{non overlapping} sampling strategy, and input sequence with length $q\!=\!3$, since training time for 1 epoch took more than 7 hours at an NVIDIA Titan RTX of 24GB. In comparison, 1 epoch of our proposed method needs only 35 minutes, mainly because recurrent layers act in a much lower dimensional space. For model \textit{ConvLSTM+Clf}, we have tried to use the \textit{argmax} values of the classifier for the \textit{heading} channel, thus having only possible existing values in that channel. In this way, the accuracy of \textit{heading} increases to more than 80\%, but the mse doubles, mainly because the difference between the wrong prediction values now is bigger than using the default prediction minimized with mse, because softmax crossentropy does not lead to closer values when not guessing the correct one, so we discard this strategy.

Due to time constraints of the challenge and the time used for the baselines to work, the rest of our experiments focus only on improving the performance for Moscow. Future work will include comparisons also for the other cities.

\begin{table}[H]
\centering
\begin{tabular}{|l|l|l|}
\hline
\multicolumn{1}{|c|}{mse (acc \textit{heading}), \#epochs} & \multicolumn{1}{c|}{\textit{ConvLSTM}} & \multicolumn{1}{c|}{\textit{ConvLSTM+Clf}} \\ \hline
Moscow                                            & 0.012643729 (0.265), 2        & 0.\textbf{012037826} (0.455), 2+2           \\ \hline
Istanbul                                          & 0.\textbf{009066001} (0.657), 3        & 0.0096280435 (0.686), 3+3         \\ \hline
Berlin                                            & 0.0071446532 (0.536), 5       & 0.\textbf{007143738} (0.418), 5+1          \\ \hline
\end{tabular}
\caption{Baseline comparison for all three cities.  We show global mse and accuracy of the heading channel together with epochs (using '+' if ConvLSTM+Clf started from last epoch in ConvLSTM).}
\label{tab:baselines}
\vskip -0.5 cm
\end{table}

In Table~\ref{tab:best_models_moscow} we show the performance of the proposed method and its variants for Moscow. The best model, in bold, is \textit{RAE\_all} (see Table~\ref{tab:moscow_best} for more details), which was fine-tuned from \textit{RAE\_not\_In} with a new skip connection from the last frame in the input sequence to the decoder, probably allowing to further refine static regions. It is worth to mention that model \textit{RAE\_Clf} is really learning to be more accurate with the \textit{heading}, future work will focus more in this model since evaluation uses mse but still, we desire that heading channel outputs only the possible values.
\begin{table}[H]
\centering
\resizebox{\columnwidth}{!}{%
\begin{tabular}{l|l|l|l|l|l|}
\cline{2-6}
                                  & \textit{ConvLSTM+Clf} & \textit{RAE\_not\_Exo}  & \textit{RAE\_not\_In}   & \textit{RAE\_all} & \textit{RAE\_Clf} \\ \hline
\multicolumn{1}{|l|}{mse}                  & 0.012037826  & 0.011873306      & 0.011875369 & \textbf{0.011816756} & 0.014442413     \\ \hline
\multicolumn{1}{|l|}{\textit{heading} acc} & 0.455        & 0.469            & 0.453       & 0.437                & \textbf{0.508}  \\ \hline
\multicolumn{1}{|l|}{epochs}               & 4            & 10               & 10+5        & 15+3                 & 44              \\ \hline
\end{tabular}
}
\caption{Comparison of results for our proposed method and its variations against the baseline in Moscow city. We show global mse and accuracy of the heading channel. Epochs are also shown with the particularity that if they are shown with format 10+5, it means that it is a fine-tuned model trained 5 epochs with loaded weights of the model at its left, that was trained 10 epochs already.}
\label{tab:best_models_moscow}
\vskip -0.5 cm
\end{table}

\begin{table}[H]
\centering
\begin{tabular}{llll}
                     & \multicolumn{3}{c}{\begin{tabular}[c]{@{}c@{}}Moscow | mse: 0.011816756 \end{tabular}} \\ \hline
\multicolumn{1}{c}{} & \multicolumn{1}{c}{\textit{speed}}                       & \textit{volume}                             & \textit{heading}                           \\ \hline
5 minutes            & 0.000095909214                                   & 0.005128963                        & 0.029793534                       \\
10 minutes           & 0.000102340266                                  & 0.0051734922                       & 0.030223647                       \\
15 minutes           & 0.00010444787                                   & 0.0052143666                       & 0.03051411                       \\ \hline
\end{tabular}
\caption{Complete mse results in Moscow city for our best model \textit{RAE\_all}. As expected, the closer the time bin ahead to predict, the better the result is.}
\label{tab:moscow_best}
\vskip -0.5 cm
\end{table}

Our proposed architecture for the city of Moscow together with the best baseline for Istanbul and Berlin achieved a mse score of 0.0098134960517874 in the challenge. We believe that using our method well trained for all cities will lead to much more better performance, which could not be tried due to time constraints.

\section{Conclusions} \label{Conclusions}

In this paper, a Recurrent Autoencoder with Skip Connections and Exogenous Variables has been proposed for Traffic Forecasting under a novel representation of traffic data, introduced by the Traffic4cast Challenge at NeurIPS 2019, that aggregated traffic data for cities into images, generating videos of traffic behavior.
Our proposed method leverages the fact that the input is a sequence of frames, aggregating all information in a lower dimensional space and generating the output sequence in a single inference. The model uses two loss functions ensuring good predictions in the embedding space and high definition images in the original space. Exogenous variables like the day of the week, time of the day and weather have been also included. Furthermore, we propose a sampling method for sequences that run in parallel to the optimization process, producing rich batches in terms of diversity at each epoch. The result reported on the Traffic4cast Challenge is a mse of 0.0098134960517874 in the test set. In future work, we will apply our method to Berlin and Istanbul and focus on the multitask method that uses the results of the classification of the \textit{heading} channel as an input for the real prediction.

\subsubsection*{Acknowledgments}
This work is supported by SEAT, S.A., and the Secretariat of Universities and Research of the Department of Economy and Knowledge of the Generalitat de Catalunya, under the Industrial Doctorate Grant 2017 DI 52. This research is also supported by the grant TIN2017-89244-R from MINECO (Ministerio de Economia, Industria y Competitividad) and the recognition 2017SGR-856 (MACDA) from AGAUR (Generalitat de Catalunya).

\end{document}